\pdfoutput=1
\documentclass[11pt]{article}

\usepackage[preprint]{acl}

\usepackage{times}
\usepackage{latexsym}

\usepackage[T1]{fontenc}

\usepackage[utf8]{inputenc}

\usepackage{microtype}

\usepackage{inconsolata}

\usepackage{graphicx}

\usepackage{multirow}
\title{Towards Efficient Large Language Models for Scientific Text: A Review}

\author{
 \textbf{Huy Quoc To},
 \textbf{Ming Liu},
 \textbf{Guangyan Huang}
\\
School of Information Technology, Deakin University, Australia
\\
\{q.to, m.liu, guangyan.huang\}@deakin.edu.au
}

\begin{document}
\maketitle
\begin{abstract}
    Large language models (LLMs) have ushered in a new era for processing complex information in various fields, including science. The increasing amount of scientific literature allows these models to acquire and understand scientific knowledge effectively, thus improving their performance in a wide range of tasks. Due to the power of LLMs, they require extremely expensive computational resources, intense amounts of data, and training time. Therefore, in recent years, researchers have proposed various methodologies to make scientific LLMs more affordable. The most well-known approaches align in two directions. It can be either focusing on the size of the models or enhancing the quality of data. To date, a comprehensive review of these two families of methods has not yet been undertaken. In this paper, we (I) summarize the current advances in the emerging abilities of LLMs into more accessible AI solutions for science, and (II) investigate the challenges and opportunities of developing affordable solutions for scientific domains using LLMs.
\end{abstract}

\section{Introduction}
Recently, the advancement of large language models (LLMs) has equipped us with the capability to address complex tasks that demand an understanding of both structure and language. The key factors that make LLMs so rapid are the huge amount of generated data and the advancement in computational architectures. With regard to scientific data itself, this domain has witnessed a constantly and rapidly increase in number of publications. For example, there were more than 2.4 million scholarly papers on ArXiv\footnote{\url{https://arxiv.org/stats/monthly_submissions}} (up to 2024) and 36 million publications on PubMeb\footnote{\url{https://www.nlm.nih.gov/bsd/medline_pubmed_production_stats.html}} (up to 2022). The exponential growth enables us to leverage the success of language models to effectively learn scientific knowledge. Recently, \cite{ho2024survey} reported that there are about 117 language models constructed for the scientific domain. Tasks such as Text Classification, Summarization, or Named-Entity Recognition are effectively handled by most of these models, which have shown impressive performance on various benchmarks. 

In order to perform sophisticated problem-solving tasks, the scientific language models are designed to have complex structures with vast scale. In particular, recent LLMs for science such as Galactica \cite{taylor2022galactica} are equipped with groundbreaking architectures. They surpass most of the evaluations on reasoning, problem solving, and knowledge understanding. However, these LLMs face inevitable drawbacks, as they require a substantial amount of resources, for example, a large-scale high-quality dataset and a high training or inference cost \cite{openai2024gpt4}. Whereas, these resources are not available in many cases, such as low-resource languages or small organizations with limited computational access. Therefore, limitations related to accessibility, cost, and adaptability pose substantial challenges to fully utilize the capabilities of scientific LLMs. In this review, we present two main contributions:
\begin{itemize}
    \item We provide a comprehensive overview of the latest developments of the application of Large Language Models (LLMs) in scientific fields. This includes discussing how LLMs have been tailored to solve complex scientific problems, and their integration into existing studies.
    \item We delve into examining the technical and economic barriers to deploying LLMs for science, exploring cost-effective strategies and innovations, and identifying opportunities for reducing expenses without compromising performance. 
\end{itemize}

\section{Related surveys}
There are few surveys on pre-trained language models (PLM) for science \cite{ho2024survey, KALYAN2022103982, wang2023pretrained} and to make LLMs more accessible \cite{wan2024efficient, xu2024survey}. Regarding scientific language models, \cite{ho2024survey} presented the first comprehensive review of scientific language models (SciLM), describing more than 110 models, evaluating their performance across various domains and tasks, and addressing future research challenges. The study analyzed six key dimensions: scope, target language models, domains, scientific texts, languages, and modalities, offering a distinctive evolutionary perspective on SciLMs over recent years. Specifically, in the biomedical sector, \cite{wang2023pretrained} reviewed the latest advancements of PLMs in the biomedical field and their applications in downstream biomedical tasks. The authors explored the motivations for PLMs in the biomedical sector, outlined key concepts, and proposed a taxonomy that classifies existing biomedical PLMs from multiple perspectives. In a different related survey by \cite{KALYAN2022103982}, researchers explored the core principles of transformer-based PLMs, such as pre-training techniques, pre-training tasks, fine-tuning strategies, and embedding types tailored to the biomedical domain. The study presented a classification system for transformer-based PLMs, evaluated all existing models, identified several challenges, and proposed possible solutions.

\par According to \cite{wan2024efficient}, while LLMs are at the forefront of the AI revolution, their impressive abilities require significant resources. As model sizes increase, the GPU hours needed for training increase exponentially, enhancing performance but also increasing costs. Furthermore, inference operations significantly add to the financial burden of running LLMs. Although enlarging the size of the model improves performance, it reduces inference throughput (increases inference latency), which poses obstacles in extending their adoption to a wider range of customers and applications affordably. The substantial resource requirements of LLMs underscore the critical necessity of devising methods that improve their efficiency. In the survey of \cite{wan2024efficient}, a fairly detailed number of approaches based on three aspects is listed: model-centric, data-centric, and frameworks. However, their survey lacks investigation on the application of listed methods in different domains. 
Additionally, the survey conducted by \cite{xu2024survey} emphasizes harnessing the capabilities of proprietary LLMs (like those from the GPT family) through knowledge distillation. This technique involves transferring the implicit 'knowledge' from proprietary models into open-source language models. The goal of these methods is to narrow the performance gap between state-of-the-art proprietary and open-source LLMs. By using the advanced features of leading proprietary models such as GPT-4 \cite{openai2024gpt4} as benchmarks, knowledge distillation aims to enhance the performance of open-source LLMs. This method resembles an experienced instructor transferring expertise to a student, with the student models adopting the performance traits of the teacher LLMs. 
\par Despite the existing surveys on making LLMs more accessible, these works presented methods and techniques primarily in a broader domain. Meanwhile, in previous reviews on scientific language models, the authors encouraged finding efficient and low-cost solutions for scientific adaptation and leveraging LLMs for science. Therefore, our review focuses on investigating recent efficient approaches for scientific LLMs and potential research directions. 
\section{Advancement in efficient LLMs for Science}
\begin{figure}[h]
\centering
    \includegraphics[scale=0.4]{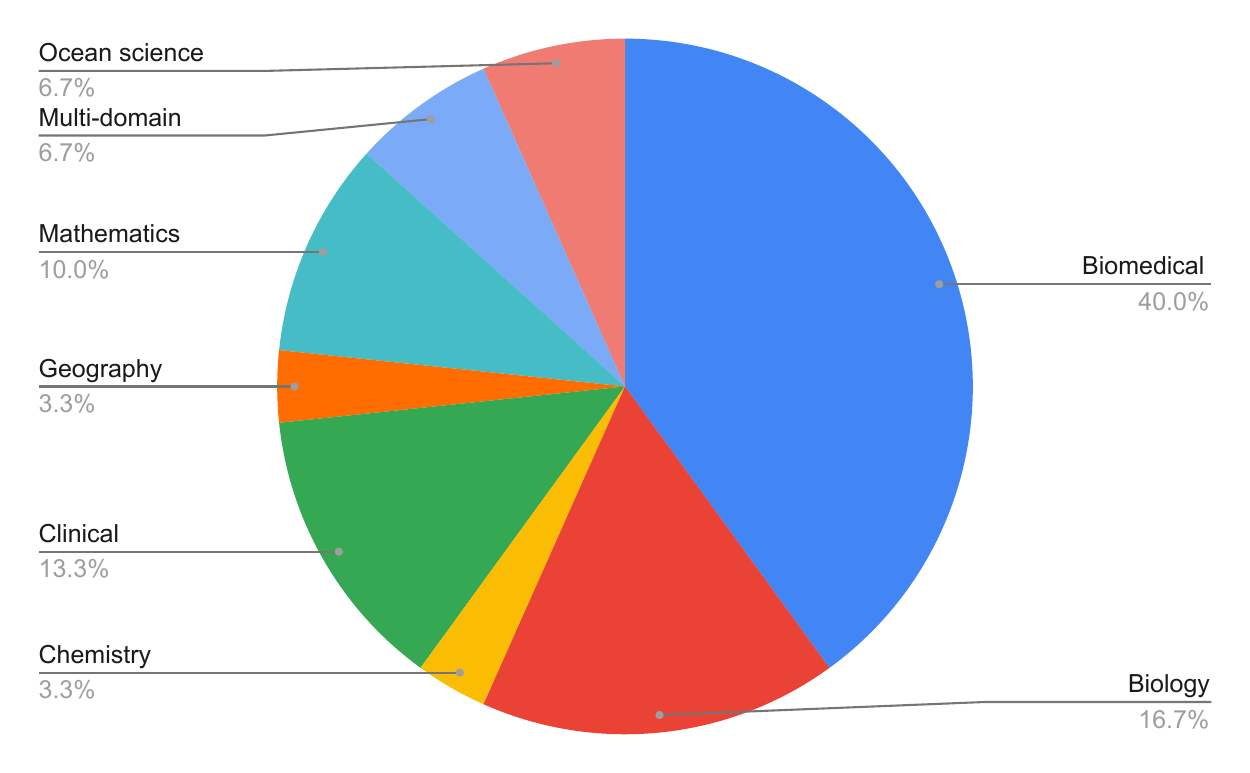}
    \caption{Distribution of efficient LLMs for Science.} 
    \label{fig:dist}
\end{figure}
This section discusses the latest developments in the application of Large Language Models (LLMs) within the scientific field. The purpose of this study is to investigate the capabilities of LLMs in scientific research. In this review, we attempt to encompass a broad range of science-related topics, including biology, biomedicine, mathematics, geoscience, ocean science, and other natural sciences. Figure \ref{fig:dist} shows the distribution of efficient methods leveraging LLMs for each scientific domain in our review.
\subsection{Biology} In the field of biology, there has been a trend towards studying increasingly large language models. Yet, the substantial computational and memory requirements for fine-tuning these models pose significant challenges for many academic laboratories and small biotechnology firms. \cite{Sledzieski2023} implemented parameter-efficient fine-tuning (PEFT) on ESM2 model \cite{esm2} to predict protein-protein interactions. Employing the PEFT technique LoRA, the model surpassed the performance of fully fine-tuned model while consuming less memory, illustrating that effective deployment of large protein language models is feasible even for groups with constrained computational resources. The study further highlighted that the efficacy of this method could be enhanced by utilizing more informative embeddings produced by LLMs. Research on the PEFT LoRA method adapted for the ESM2 model was also conducted in \cite{Zeng2023} focusing on signal peptides (SP) prediction. Other PEFT techniques such as Adapter Tuning and Prompt Tuning were also explored. It was noted that Prompt Tuning underperformed compared to other previous models, likely due to the size of the model. While Adapter Tuning improved performance, it required a considerably larger number of training parameters relative to LoRA. Future enhancements were suggested, including combining PEFT techniques to improve interpretability for identifying SP-related motifs and integrating structure-aware language models to include protein structural data. Another PEFT approach, adaptive LoRA (AdaLoRA), was utilized in the study by \cite{zhan2023parameterefficient}. This study introduced AdaLoRa with random sampling (AdaLoRA-RS) on OPT-350M to enhance the understanding of genomic language complexities. When compared to other models, DNABERT and Nucleotide Transformer, AdaLoRA+RS demonstrated performance on par with fully fine-tuned models across 13 genomic datasets while using less than 2\% of the training parameters. The experimental findings further showed that pre-trained language models such as OPT-125M outperformed the specialized DNA model HR-500M, utilizing only 25\% of the parameters. 
\subsection{Biomedical domain} The progress of LLMs has significantly influenced biomedical research. In the last ten years, extensive unlabelled datasets like PubMed, PMC, MIMIC, and ScienceDirect have been made accessible in the biomedical field. Models such as GPT-4 and Med-PaLM 2 have demonstrated outstanding performance in a variety of biomedical NLP tasks. Nonetheless, these models, with their vast number of parameters, are costly in terms of computational resources, necessitate data transmission over the Internet, and are trained on proprietary datasets.
\par In 2022, \cite{LI2022346} leveraged this unlabelled information to introduce BioKnowPrompt, a prompt-tuning PLM framework tailored for extracting relationships from biomedical texts. Additionally, prompting can be challenging for certain phenomena and may struggle with highly imbalanced training data. Follow up, with the introduction of ChatGPT and GPT-4 \cite{openai2024gpt4}, many researches have leveraged its power for data augmentation. \cite{zhang-etal-2023-huatuogpt} created HuatuoGPT based on LlaMa model, employing both refined data from ChatGPT and data from doctors for health consultations. This model is superior at producing patient-friendly and doctor-like responses and outperformed existing medical open-source LLMs. DoctorGLM presented by \cite{xiong2023doctorglm} also used the data generated by ChatGPT for medical dialogues in Chinese. The training process of DoctorGLM can handle a considerable number of question-answer pairs per hour per GPU, with a relatively low cost per training session. Furthermore, the inference operations of DoctorGLM demand minimal GPU memory, enabling execution on standard consumer hardware, thus making it accessible for numerous research facilities and healthcare centers. They also mentioned that the model can be deployed on even more affordable GPU when applying PEFT method such as LoRA. The superiority of GPT-4 also demonstrated by \cite{Hsueh2023gpt4}. The authors succeeded in using prompt engineering for ChatGPT(GPT-4) to generate answers for biomedical questions. Although their method outperformed the fine-tuned BioBERT model, they discussed that there were rooms for improvements such as determining key information before prompting. \cite{bolton2024biomedlm} presented BioMedLM, a GPT-style autoregressive model with 2.7 billion parameters, trained solely on PubMed abstracts and full articles. Their findings emphasize that smaller models can offer transparent, privacy-preserving, cost-effective, and eco-friendly solutions for biomedical applications. Another approach using GPT-3.5 for biomedical purposes was presented by \cite{bao2023discmedllm}. The researchers employed GPT-3.5 to extract medical knowledge triples from a knowledge graph through a department-focused method based on patient query patterns from real-world consultations, producing 50,000 samples. Additionally, \cite{liu2023deidgpt} addressed the issue of securely managing medical data in the modern digital age, where confidentiality is a major concern. Utilizing advancements in large language models such as ChatGPT and GPT-4, the researchers introduced DeID-GPT, an innovative framework designed to automatically identify and mask personal information in medical texts. Their method not only achieved high accuracy in maintaining text integrity but also set a new standard for the application of LLMs in healthcare settings focused on privacy protection.

\par While proprietary LLMs are usually huge, untrainable and their architecture are unclear, researchers adapt instruction-tuning technique to open-source smaller LLMs for solving biomedical problems. \cite{wu2023pmcllama} systematically adapted the open-source general LLM, LLaMA, for biomedical tasks by injecting domain-specific data and instruction-tuning tailored to medical contexts. The PCM-LLaMA model, an open-source language model designed for medical purposes, demonstrates superior results on various medical benchmarks, surpassing both ChatGPT and LLaMA-2 while utilizing considerably fewer parameters. Additionally, \cite{luo2023biomedgpt} presented BioMedGPT, a multi-modal generative pre-trained model tailored for biomedical applications. The design of BioMedGPT highlights the critical importance of knowledge distillation in bridging complex biological information with natural language, enabling substantial advancements in the discovery of drugs and therapeutic targets. \cite{PENG2024104630} conducted comparison on GatorTron using soft-prompting in various configurations. The study revealed that soft prompting surpassed hard prompting, unfrozen Large Language Models (LLMs) display robust few-shot learning abilities and adaptability across different institutions, using frozen LLMs reduces computational costs to between 2.5\% and 6\% relative to earlier methods that utilized unfrozen LLMs, while still attaining optimal outcomes with large-scale unfrozen LLMs. To enhance performance and generalizability beyond traditional benchmarks, \cite{zhang2024alpacareinstructiontuned} introduced MedInstruct-52k, a diverse dataset generated with GPT-4 and ChatGPT. Fine-tuning LLaMA-series models on this dataset resulted in AlpaCare, which outperformed previous medical LLMs by up to 38.1\% in medical instruction-following tasks and showed consistent improvements in general domain benchmarks based on human evaluations. Parameter-efficient fine-tuning is also an effective approach to reduce the training time and cost when performance domain adaptation. \cite{han2023medalpaca} utilized the LLaMA foundation models with 7 billion and 13 billion parameters, fine-tuning them over five epochs with learning rates specifically adjusted for each model variant. They applied Low-Rank Adaptation (LoRA) to improve efficiency by lowering GPU memory usage and reducing training time. Additionally, the author incorporated 8-bit matrix multiplication to further decrease computational requirements, making it more feasible to deploy these models in medical applications with strict resource limitations.

\subsection{Clinical domain} The clinical domain has also experienced a transition from traditional pre-trained language models to the effective use of LLMs. \cite{gema2024parameterefficient} introduced a two-step PEFT framework based on the LLaMA model, which was evaluated within the clinical domain. This framework integrates a specialized PEFT adapter layer for clinical domain adaptation with another adapter for downstream tasks. It was tested on various datasets for clinical outcome prediction and compared to language models trained specifically for clinical purposes. This research is the first to propose a comprehensive empirical analysis of the interaction between PEFT techniques and domain adaptation in the crucial real-world setting of clinical applications. \cite{GOSWAMI2024111531} investigated the utility of prompt engineering and parameter-efficient fine-tuning for summarizing hospital discharge summary (HDS) articles. The goal was to ensure these models correctly interpret medical terminology and contexts, generate brief summaries, and extract key themes. The research employed LLaMA-2 as the base model and fine-tuned it using QLoRA (Quantized Low-Rank Adapters) to reduce memory consumption while maintaining data quality. Chinese patent medicine (CPM), an essential part of traditional Chinese medicine (TCM) utilizing Chinese herbs, was studied by \cite{liuchatglm2024} with LLMs. The researchers created the first CPM instructions (CPMI) dataset and fine-tuned the ChatGLM-6B base model, resulting in CPMI-ChatGLM. They employed parameter-efficient fine-tuning with consumer-grade graphics cards and investigated LoRA, P-Tuning v2, along with various data scales and configurations. Comparative experiments with similar-size LLMs demonstrated the leading performance of CPMI-ChatGLM in recommending CPM, highlighting its potential for clinical support and data analysis in TCM research.

\subsection{Mathematics} Large language models such as GPT-4 have shown remarkable abilities in complex mathematical reasoning. However, open-source models are generally pre-trained on extensive internet data without specific tuning for mathematical tasks. To address this gap, \cite{luo2023wizardmath} introduced WizardMath, which improves mathematical reasoning in LLaMa-2 using Reinforcement Learning from Evol-Instruct Feedback (RLEIF). WizardMath outperformed ChatGPT-3.5, Claude Instant-1, PaLM-2, and Minerva on GSM8k, as well as Text-davinci-002, PaLM-1, and GPT-3 on MATH, demonstrating the effectiveness of RLEIF. Building on this groundwork, \cite{yue2023mammoth} presented MAmmoTH, a series of open-source LLMs designed specifically for mathematics. MAmmoTH-7B achieved a 33\% accuracy rate on MATH, exceeding WizardMath-7B by 23\%, highlighting the significance of diverse problem coverage and hybrid rationales in the development of advanced mathematical models.

Additionally, \cite{gou2024tora} presented TORA, integrating natural language reasoning with external computational tools like computation libraries and symbolic solvers to tackle challenging mathematical problems. TORA models significantly outperformed existing open-source models on ten mathematical reasoning datasets, achieving average improvements of 13\%-19\%. TORA-7B achieved 44.6\% accuracy on the competition-level MATH dataset, outperforming WizardMath-70B by 22\% absolute, demonstrating the effectiveness of integrating computational tools with language models for mathematical problem-solving.
\subsection{Geoscience} In geoscience, \cite{deng2023k2} unveiled K2, the pioneering LLM specifically designed for geoscience applications. The researchers created essential resources to advance LLM research in geoscience, such as GeoSignal, the first dataset for geoscience instruction tuning, and GeoBench, the first benchmark for evaluating LLMs in this field. Their study described the adaptation of a pre-trained general-domain LLM, namely the LLaMA-7B model, to the geoscience domain by further training it on a 5.5 billion token corpus of geoscience texts and fine-tuning it with supervised data from GeoSignal. Additionally, the authors outlined a protocol for efficiently collecting and constructing domain-specific supervised data, even with limited resources. The experimental results on GeoBench demonstrated the effectiveness of their approach and datasets in enhancing the understanding and application of geoscience knowledge, representing a significant breakthrough in the integration of LLMs into geoscientific research and practice. 
\subsection{Chemistry} In the effort to improve crystal property prediction, recent research has focused on using textual descriptions of crystal structures. Traditional methods primarily use graph neural networks (GNNs) to model these structures \cite{huang2024adagnn, ruff2023connectivity, yan2024complete}, but they often struggle with the complex interactions between atoms and molecules. An innovative approach by \cite{rubungo2024llmprop} involves creating a benchmark dataset named TextEdge, which provides comprehensive text descriptions of crystal structures along with their properties. Furthermore, the authors propose LLM-Prop, a novel method employing large language models (LLMs) to predict the physical and electronic properties of crystals based on their textual descriptions. Remarkably, it outperforms a domain-specific fine-tuned BERT model, MatBERT, despite having significantly fewer parameters.

\subsection{Ocean Science} Ocean science, essential for comprehending the vast reservoirs of life and biodiversity that cover over 70\% of our planet, has not yet fully reaped the benefits of advancements in large language models (LLMs). Although LLMs have been successful in various domains, they often fall short in addressing the specialized requirements of oceanographers due to the complexity and richness of ocean data. To bridge this gap, \cite{bi2024oceangpt} introduced OCEANGPT, the first LLM specifically designed for ocean science. Extensive experiments revealed that OCEANGPT not only exhibited a high level of expertise in ocean science but also demonstrated initial capabilities in embodied intelligence for ocean technology. Additionally, \cite{zheng2023marinegpt} introduced MarineGPT, the first vision-language model specifically crafted for the marine domain. MarineGPT, developed using the Marine-5M dataset containing over 5 million marine image-text pairs, aimed to make ocean knowledge more accessible and enhance marine vision and language alignment, addressing the shortcomings of existing general-purpose MLLMs in understanding and responding to domain-specific intents.
\subsection{Multi-scientific domains} In their respective studies, \cite{xie2023darwin} introduced DARWIN, a series of tailored LLMs optimized specifically for scientific disciplines such as material science, chemistry, and physics. Built upon the foundational LLaMA-7B model, DARWIN achieved significant advances in automating the generation of scientific text instruction, thus improving its performance in various scientific tasks and reducing the dependency on closed-source LLMs.
Similarly, \cite{zhang2024sciglm} presented SciGLM, a suite of scientific language models designed for college-level scientific reasoning. Using a self-reflective instruction annotation framework, SciGLM addressed data scarcity challenges in the science domain by improving both base models like ChatGLM3-6B-Base by 4.87\% and larger-scale models by 2.67\%. This approach enhances the model's ability to conduct diverse scientific discovery tasks while preserving its language understanding capabilities.
\begin{table*}[t]
\centering
\begin{tabular}{|cc|p{0.6\linewidth}|}
\hline
\multicolumn{2}{|c|}{\textbf{Methods}}                                             & \textbf{Models}                                                                                         \\ \hline
\multicolumn{2}{|c|}{\textbf{Efficient Fine-tuning}}                              & ESM2-LoRA \cite{Sledzieski2023}, PEFT-SP\cite{Zeng2023}, AdaLoRA+RS \cite{zhan2023parameterefficient}, BioMedGPT \cite{luo2023biomedgpt}, MedAlpaca \cite{han2023medalpaca},  Clinical LLaMA-LoRA \cite{gema2024parameterefficient}, LLaMa-QLoRA \cite{GOSWAMI2024111531}, CPMI-ChatGLM \cite{liuchatglm2024}          \\ \hline
\multicolumn{2}{|c|}{\textbf{Instruction Tuning}}                                   & BioKnowPrompt \cite{LI2022346}, NCU-IISR \cite{Hsueh2023gpt4}, Alpacare \cite{zhang2024alpacareinstructiontuned}, GatorTron \cite{PENG2024104630}, K2 \cite{deng2023k2}, WizardMath \cite{luo2023wizardmath}, MAmmoTH \cite{yue2023mammoth}, TORA \cite{gou2024tora}, OCEANGPT \cite{bi2024oceangpt}, MarineGPT \cite{zheng2023marinegpt}, SciGLM \cite{zhang2024sciglm} \\ \hline
\multicolumn{1}{|c|}{\multirow{2}{*}{\textbf{Knowledge distillation}}} & Black box & HuatouGPT \cite{zhang-etal-2023-huatuogpt}, DoctorGLM \cite{xiong2023doctorglm}, DISC-MedLLM \cite{bao2023discmedllm}, DeID-GPT \cite{liu2023deidgpt}                                                         \\ \cline{2-3} 
\multicolumn{1}{|c|}{}                                        & White box & BioMedLM \cite{bolton2024biomedlm}, PCM-LLaMA \cite{wu2023pmcllama}, LLM-Prop \cite{rubungo2024llmprop}, DARWIN \cite{xie2023darwin}                                               \\ \hline
\end{tabular}
\caption{Summary of previous work on efficient LLMs for science.}
\label{tab:summary}
\end{table*}
\section{Challenges and future directions}
Current studies on the application of LLMs in science have made significant progress.  We summarize the existing methods and scientific LLMs in Table \ref{tab:summary}. Most of these studies have initially harnessed the power of LLMs to address problems in scientific fields such as biology and biomedicine. However, many issues remain unresolved. This section will present some research gaps with potential for further exploration.
\subsection{Data Collection}
\paragraph{Challenges} The lack of labeled data is a common issue faced by researchers when training language models in various scientific fields. Despite the abundance of unlabeled scientific data, it is not utilized efficiently to train language models. \cite{ho2024survey} summarized that among 117 language models for scientific fields, most previous work focused on the biomedical domain, with more than 87\% pre-trained language models in this area. The author also noted that these language models typically have fewer than 1 billion parameters (e.g., BERT-based models) and do not leverage open-source LLMs. \textbf{This leads to the issue of underutilization of unlabeled data in various scientific domains.} The process of gathering high-quality labeled data for training models is time-consuming and labor-intensive. 
\paragraph{Potential directions} Current approaches like active learning for small language models (SLMs) and in-context learning for large language models (LLMs) have partially alleviated the shortage of labeled data, yet they still depend heavily on human involvement. \cite{xiao-etal-2023-freeal} tackled this problem by proposing FreeAL, a collaborative learning framework where an LLM serves as an active annotator and an SLM filters high-quality in-context samples for label refinement. Comprehensive experiments on eight benchmark datasets demonstrated that FreeAL significantly enhanced zero-shot performance for both SLMs and LLMs without human supervision. \cite{zhang2023llmaaa} introduced LLMaAA, which uses LLMs as annotators in an active learning loop to efficiently select data for annotation, demonstrating superior performance in named entity recognition and relation extraction tasks with fewer annotated examples. \cite{huang2024selective} tackled the challenge of high quality annotations under limited budgets with SANT, a selective annotation framework utilizing error-aware triage and bi-weighting mechanisms, setting a new benchmark for triage-based annotation studies.

\subsection{Data Selection}
\paragraph{Challenges} Determining the optimal data volume crucial for maximizing the effectiveness of Large Language Models (LLMs) remains a persistent challenge, necessitating further research to establish clear guidelines. Additionally, developing robust methodologies to filter out low-quality data continues to be an ongoing concern in leveraging LLMs effectively.
\paragraph{Potential Directions} In general domain, \cite{zhou2023lima} proposed that a minimum of 1000 well-curated, high-quality data samples could be sufficient to align LLMs, as pre-training already provides essential knowledge. \cite{chen2024alpagasus} introduced a new data selection method using a robust LLM such as ChatGPT to independently filter out low-quality data. They developed AlpaGasus, a model refined with just 9,000 high-quality samples from the initial dataset. More recently, \cite{li2024superfiltering} presented Superfiltering, which used smaller models such as GPT-2 to extract a high-quality subset from a dataset. Despite these advancements, the challenges of selecting optimal data for refining LLMs and determining the necessary data volume persist because of the abundance of unlabeled scientific data.

\subsection{Utilizing multiple LLMs}
\paragraph{Challenges} The majority of current models originate from a single LLM, yet it is commonly recognized that models trained with diverse data sources possess distinct advantages. Consequently, the question arises: \textbf{Can knowledge from multiple LLMs be integrated into a single smaller model?}
\paragraph{Potential directions} To develop a "BabyLM," \cite{timiryasov-tastet-2023-baby} trained a combination of GPT-2 and small LLaMA models on the 10M-word BabyLM dataset, subsequently distilling this ensemble into a compact, 58M-parameter LLaMA model. The distilled model surpassed both its original models and a comparable model trained without distillation, indicating that distillation can preserve and even enhance the performance of teacher models, especially on small datasets. \cite{wan2024knowledge} later introduced 'knowledge fusion' to integrate the strengths of multiple LLMs, testing their method with Llama-2, MPT, and OpenLLaMA across various benchmarks. This approach enhanced the target model's capabilities in reasoning, common sense, and code generation. Furthermore, \cite{chen2024magdi} proposed MAGDI to improve reasoning in small models by distilling interactions between multiple large LLMs using Multi-Agent Interaction Graphs (MAGs). MAGDI outperformed conventional distillation techniques and boosted reasoning and efficiency in smaller models. Despite these progressions, the scientific community still lacks extensive research on integrating knowledge from multiple LLMs. 
\subsection{Addressing Catastrophic Forgetting}
\paragraph{Challenges} Prior studies have investigated optimizing LLMs to enhance their directive-following and knowledge transfer abilities, leveraging advancements in LLM technology. However, \textbf{persistent optimization with specific datasets can lead to catastrophic forgetting}. 
\paragraph{Potential directions} In the scientific domain, \cite{yue2023mammoth} introduced MAmmoTh, an ensemble of open-source LLMs designed to tackle mathematical challenges using the MathInstruct dataset, overcoming catastrophic forgetting seen in prior models like WizardMath \cite{luo2023wizardmath}. Meanwhile, continual learning (CL) research focuses on dynamically enhancing models while preserving prior knowledge. Approaches like Lifelong-MoE \cite{chen2023lifelong}, CITB \cite{zhang-etal-2023-citb}, and DCL \cite{zeng2023continual} employ techniques such as adding experts, regularization, task distribution modeling, and knowledge distillation to combat catastrophic forgetting. Nevertheless, preserving the initial model's abilities and transferring knowledge across different domains continue to pose significant challenges.
\subsection{Multimodality}
\paragraph{Challenges} Within the scientific field, there is increasing enthusiasm for multi-modal models \cite{ho2024survey}, which are created by further training mono-modal or multi-modal models from general domains, capitalizing on their robust performance. Nonetheless, numerous challenges remain. The scientific domain often suffers from a scarcity of data compared to general domains, complicating the process of effectively training or fine-tuning multi-modal language models. \textbf{Integrating this multi-modal information into scientific language models is essential for research progress.}
\paragraph{Potential directions} Several research efforts are dedicated to creating adapters that transform non-language data to be interpreted within the same embedding space as language \cite{dai2023instructblip, zhu2023minigpt4}. These frameworks are designed to manage non-language information while maintaining the strong problem-solving abilities of LLMs. Although proprietary LLMs such as GPT-4 can handle various scientific data types, utilizing these models demands substantial resources. Consequently, it is advisable to develop efficient strategies to enhance the accessibility of LLMs and incorporate multimodality in scientific domains, thereby unlocking the full potential of multi-modal models in the scientific field.
\subsection{Further reduce the cost}
\paragraph{Challenges} Despite the impressive capabilities of modern LLMs, their substantial resource demands highlight \textbf{the critical need for effective solutions to address these challenges}. Based on Table \ref{tab:summary}, in previous work within the scientific domain, common ways to reduce costs have included Instruction Tuning and Efficient Fine-Tuning. Continued research and development in other methodologies are crucial to making LLMs more accessible and sustainable.
\paragraph{Potential directions} In other fields, numerous efficient methods have been explored, including Quantization \cite{frantar2023optimal, kim2023finequant, tao-etal-2022-compression}, Parameter Pruning \cite{ma2023llmpruner, zhang2023loraprune}, and Memory Efficient Fine-Tuning \cite{dettmers2023qlora, malladi2023finetuning}. The challenge of further reducing the cost of LLMs remains unresolved. For example, Memory Efficient Fine-Tuning approaches, such as QLoRA \cite{GOSWAMI2024111531}, which enhances memory efficiency during fine-tuning, also present viable solutions. 
\section{Conclusion}
The swift progress of large language models (LLMs) has greatly improved our capability to tackle intricate tasks that demand profound linguistic and structural comprehension. The growth of scientific data has enabled effective learning of scientific knowledge through LLMs. However, despite their impressive performance in tasks like reasoning and problem-solving, these models remain resource-intensive and often inaccessible to smaller organizations and low-resource languages. Our review highlighted various cost-effective techniques for utilizing LLMs in scientific domains. We address the challenges in fully harness the potential of LLMs for science and ensure their broader accessibility and applicability in scientific research.
\section{Limitations}
Our work is based on results and suggestions of as many papers as possible we can find. We also mostly emphasize text-based scientific information, setting aside other forms such as images, videos, audio, and structured knowledge like knowledge graphs (KGs) and databases for future consideration. Our review primarily highlights the most recent advancements in the last three years, specifically from 2023 and 2024. However, our review may hold a potential of missed out some the most recent studies. We leave this as future improvements. Moreover, due to space limitations, we provide only concise summaries of the reviewed methods.
\bibliography{main}




\end{document}